# Artificial Intelligence in Traffic Systems

Ritwik Raj Saxena[1*]

*1. Department of Computer Science, University of Minnesota, Duluth Campus, Duluth, 55813.*

## Abstract

Background and Motivation: Traffic management is a pressing challenge in modern societies. The population of humans is amassing at a substantial pace, and along with that snowballs the expanse of urban areas as well as the number of private as well as public vehicles. This makes it increasingly more complicated to surveil and manage all transport modalities at the same time and at tenable costs. Also, with an expanding number of automobiles, vehicular congestion, a growing number of choke points and increased instances of on-road disruption collectively become rapidly burgeoning traffic management problems, especially in urban areas. These issues pose an exceedingly complex challenge for metropolitan communities, leading to financial losses, delays in the delivery of emergency services to people, environmental pollution, and a reduced quality of life. Artificial Intelligence (AI) has stood out as a potent instrument towards resolving such questions. It has the capacity to augment traffic flow, accentuate transportation effectiveness, and raise the reassurance levels of passengers, commuters, as well as pedestrians. This article attempts to elucidate a myriad of applications of AI in the province of transportation management. It examines its potential to revolutionize urban transportation.

Objective: Existing research on AI-based traffic management systems, utilizing techniques such as fuzzy logic, reinforcement learning, deep neural networks, and evolutionary algorithms, demonstrates the potential of AI to transform the traffic landscape. This article endeavors to review the topics where AI and traffic management intersect. It comprises areas like AI-powered traffic signal control systems, automatic distance and velocity recognition (for instance, in autonomous vehicles, hereafter AVs), smart parking systems, and Intelligent Traffic Management Systems (ITMS), which use data captured in real-time to keep track of traffic conditions, and traffic-related law enforcement and surveillance using AI.

Discussion: AI applications in traffic management cover a wide range of spheres. The spheres comprise, inter alia, streamlining traffic signal timings, predicting traffic bottlenecks in specific areas, detecting potential accidents and road hazards, managing incidents accurately, advancing public transportation systems, development of innovative driver assistance systems, and minimizing environmental impact through simplified routes and reduced emissions. The benefits of AI in traffic management are also diverse. They comprise improved management of traffic data, sounder route decision automation, easier and speedier identification and resolution of vehicular issues through monitoring the condition of individual vehicles, decreased traffic snarls and mishaps, superior resource utilization, alleviated stress of traffic management manpower, greater on-road safety, and better emergency response time.

Challenges: The article acknowledges the challenges associated with implementation of AI-activated transportation management systems, such as acquisition of reliable data, concerns associated with data privacy, computational costs, and cybersecurity threats like adversarial attacks. It highlights the need for high-quality, real-time data to train and maintain AI models. There are additional challenges which are related to the integration of AI with existing traffic management infrastructure. Redressing these challenges would ensure that the public trust in such systems is maintained. Further, the existence of ethical considerations around bias in AI algorithms, particularly Natural Language Processing (NLP) models, including gender insensitivity of AI models, creates another potential hurdle.

<u>Conclusion and Future Work</u>: AI has the potential to engender a quantum shift in traffic management by bringing about smarter and more resilient transportation systems. This article underlines the need to overcome existing challenges in the operation of AI-regulated traffic management systems which will ensure their seamless performance. This will serve to perfect the on-road experience of people and bring advancement in their quality of life. The future of AI in traffic management is studded by potential applications in the field like AI-maneuvered traffic forecasting, real-time traffic updates, and personalized travel assistance. The future AI-driven traffic management systems are projected to be more comprehensive in terms of applications, powerful, holistic, ethical and inclusive, environmentally sustainable, robust, maintainable, more easily operable and feasible in terms of costs connected with time and resources utilization.

# Introduction, Background and Motivation

The flourishing field of AI has penetrated various domains, and the field of traffic systems and transportation is no exception. Metropolitan areas are grappling with exceeding complexity in traffic management. In this backdrop, the implementation of intelligent systems materializes as a promising solution. Advanced algorithms and modern methods of data analytics have been used to power intelligent systems. These systems brandish the potential to revolutionize how we navigate roads and systematize the flow of vehicles. This article delves into the intersection of AI and traffic management. It endeavors to explore the transformative applications and potential benefits of this technology [1][2][3][4][5][6][7].

*Definition of AI in traffic management*

AI in traffic management refers to the application and strategic deployment of AI and associated technologies to adjust, augment and enrich the flow, safety, economy, efficiency, and ecological sustainability of traffic and to automate transportation systems. AI algorithms are built to analyze and process huge quantities of data that is obtained from a large variety of sources, including sensors, cameras, and historical traffic patterns, to make intelligent decisions and automate traffic management tasks [8]. AI, especially machine learning algorithms, are poised to learn from thus data and improve performance of traffic systems over. The algorithms help AI-grounded traffic systems to dynamically adapt to changing traffic conditions over a large duration as well as in real-time. These systems also allow for proactive traffic management by being able to forecast future traffic trends based on past data and current observations, which are carried out using advanced sensors or whose insights are fed to the models by humans.

A significant component of AI in traffic management systems is their utility in developing strategies to minimize traffic crowding, especially since bunching up of vehicular traffic is a significant transportation problem in the current era, plaguing the roads and arteries not only of major urban centers but also of smaller towns. AI-powered systems prioritize the passage of, inter alia, ambulances, police vehicles, fire engines, for timely and effectual delivery of essential services and emergency response. This is carried out by dynamic adjustment of traffic signals and rerouting traffic based on real-time updates on road conditions [9]. One of the earliest applications of AI in traffic management was automated traffic signal control. Incident management and parking guidance are other manifestations of automation within AI in traffic management.

Adjusting traffic signal timings based on real-time traffic conditions is an example of AI in traffic management. This process is usually powered by advanced sensors and fuzzy systems, but can also involve neural networks [10]. It helps maintain a smooth traffic flow. Furthermore, being fed with real-time data on accidents, road closures, and other incidents, AI systems can be used to implement appropriate responses concomitantly. AI-centered systems are leveraged to provide drivers with real-time route recommendations in order to help them avoid areas with vehicular huddling, thereby minimizing travel time [11]. These systems use real-time data on parking availability to direct drivers to available parking spaces. This is termed AI-based parking management [12].

Perhaps the most well-known application of automation and AI in traffic management and transportation systems is self-driving cars. The integration of AI into self-driving cars enables them to navigate roads safely and efficiently. These vehicles, equipped with sophisticated AI algorithms, are capable of navigating complex road environments without human intervention. The integration of advanced sensors, such as LiDAR and high-definition cameras, in self-driving vehicles, enables such vehicles to efficiently perceive their surroundings, make driving-associated decisions in real-time, and execute the required on-the-road maneuvers with precision and efficiency [13]. Self-driving cars stand as an epitome of ultra-modern advances in transportation and improved accessibility for all.

*Overview of Traditional Traffic Management*

Traditional traffic management systems and AI-operated traffic systems represent two distinct approaches to addressing the challenges of urban transportation [14]. Traditional methods, many of which continue to be widely utilized, have served their purpose for several years. Innovative AI-driven transportation solutions involve significant advantages in terms of efficacy and resilience.

Conventional traffic management systems are commonly termed rule-based traffic management systems [15]. They are rooted in deterministic, formularized paradigms. They employ predefined algorithms, inert traffic models, and fixed parameters to regulate traffic. These systems can be computational or manual in nature, or a combination of the two, but are not necessarily considered intelligent. They operate within a framework of predetermined conditions as they rely on static procedures to govern the timing and sequencing of traffic signals. They have shown some effectualness in controlled environments. Conversely, they encounter limitations when applied to dynamic and arbitrary traffic patterns. These systems struggle to adapt to evolving traffic dynamics and to rapidly changing conditions, such as unexpected accidents, the presence of construction zones, road repair work, sudden weather events and other special happenings, in real-time. Consequently, their ability to minimize vehicular clustering is compromised. This leads to suboptimal performance.

Rule-based systems often require human operators to monitor traffic conditions and make adjustments [16]. Human operators are usually tasked with monitoring traffic conditions, identifying and communicating on-road anomalies and disruptions, and making manual adjustments to traffic signal timings and other control parameters. This high degree of human involvement is time-consuming as well as subjective and prone to errors. These systems have limited scalability. They also involve much manpower. These systems are characterized by disadvantageous response times, inefficiencies in decision-making, and heightened operational costs.

*Summary of the Challenges Associated with Traditional Traffic Management*

Traditional traffic management systems rely heavily on static infrastructure and manual processes. They face several challenges while being applied in modern urban environments (Figure 1). Suffering from ineffective resource allocation, conventional traffic systems lack the capacity to fully utilize available assets such as road infrastructure. These systems do not effectively allocate road space.

**Ineffective resource allocation and underutilization of assets like road infrastructure.**

**Inefficient apportionment of road space and lack of coordination with public transportation.**

**Unproductive utilization of traffic signals and parking spaces.**

**Lack of flexibility, limited data utilization, subjective intervention, and a deficit in active coordination.**

**Difficulty in integrating with other transportation systems like PRT, trams, streetcars, light rail transit systems, metro rail transit systems, railways, buses and airlines.**

**Rising costs of maintaining and upgrading aging infrastructure.**

**Higher environmental impact due to outdated methods and excessive vehicle use.**

Figure 1: A Digest of the Challenges Associated with Traditional Traffic Management

They are not able to efficaciously coordinate with public transportation networks. This leads to delays in reaching destinations, missed connections for passengers, overcrowding on roads and in public transport, and decreased ridership in public transport. In these systems, there is inefficiency in utilizing traffic signals and parking spaces. They also fail to address the multidimensional challenges inherent in modern traffic management environments. This happens due to a lack of flexibility, limited data utilization, subjective intervention, and a dearth of coordination.

These systems are exceedingly being pointed out for their lack of easy amenability to be integrated with other transportation systems like personal rapid transit (PRT) systems, trams, streetcars, light rail transit systems, metro rail transit systems, railways, buses and airlines. They also struggle to integrate parking systems and other infrastructure components. This lack of coordination prevents cities from managing traffic holistically. It further limits the general effectiveness of traffic control efforts (Figure 1).

Rising costs establish another challenge. Maintaining and upgrading aging infrastructure is expensive, especially as cities expand and traffic volumes grow. Lastly, traditional systems often have a higher environmental impact, as they rely on outdated methods that usually contribute to pollution. Excessive vehicle use for traffic control and emergency response exacerbates environmental damage. This highlights the need for more sustainable solutions like ITS.

*An Outline of AI-Enhanced Traffic Management Systems*

AI-fueled traffic management systems offer a strong contrast to rule-based traffic management systems. They are data-driven in nature, and can utilize real-time data from sensors, cameras, and connected vehicles to make informed decisions. They leverage powerful machine learning, reinforcement learning, and other intelligent models. They autonomously analyze and interpret historical data which is combined with data collected in real-time. This involves vast datasets. They can discern patterns and anomalies that elude the conventional rule-based approach. This transition to automated decision-making augments operational efficacy. AI-oriented traffic systems are a more appropriate embodiment of the synergy between human expertise and AI-driven technologies, revolutionizing the landscape of traffic management with unparalleled sophistication and adaptability.

AI-infused traffic systems adeptly forecast traffic conditions such as peak traffic density during rush hours, unexpected traffic surges and bottlenecks due to extraordinary events such as road accidents, and fluctuations in traffic flow patterns based on historical data analysis. These systems can also preemptively identify potential disruptions such as unforeseen road closures, construction zones impacting traffic flow, adverse weather conditions afflicting road safety, and unanticipated changes in traffic volume due to emergencies like tornadoes. These systems involve the amalgamation of diverse data sources and cutting-edge computational models. By incorporating a combination of historical traffic patterns, weather data, and real-time sensor readings, these systems extrapolate future traffic scenarios with remarkable precision. They deploy sophisticated machine learning algorithms such as convolutional neural networks (CNNs) (for image recognition in traffic sign classification, real-time position understanding for AVs etc.) [17]. They also leverage advanced time series-based predictive modeling techniques like recurrent neural networks (RNNs) [15] and long short-term memory (LSTM) networks [18], which enables these systems to capture temporal dependencies and subtle fluctuations in traffic flow. This also empowers them to proactively adjust traffic signal timings, reroute vehicles dynamically, and strengthen traffic efficiency. Many of these systems encompass the fusion of predictive analytics and big data processing. They ease urban mobility and pave the way for a smarter and more responsive transportation ecosystem characterized by anticipatory decision-making and seamless traffic flow management.

AI-triggered traffic systems make informed decisions to alleviate resource allocation. These systems feature a remarkable capacity to dynamically adjust traffic signal timings at intersections, improve lane assignments based on traffic flow patterns and allocate resources such as traffic enforcement personnel and emergency services in a responsive and adaptive manner. They depict the ability of adaptive signal control [19] by continuously adjusting traffic signals and other parameters for the better based on evolving traffic patterns. They not only boost the operational efficiency of transportation networks but also mitigate traffic snarls, palliate travel times, bolster safety on roadways, and augment sustainability and inclusivity of transportation systems.

AI-based traffic management systems serve as a cornerstone in revolutionizing urban safety paradigms by proactively identifying and minimizing potential hazards on roadways. They enable swift responses through automated alerts, dynamic rerouting strategies, and facilitated emergency service dispatches. They engender the seamless coordination of connected vehicle technologies [20], roadside sensors, and centralized AI algorithms which facilitate the early detection of hazardous conditions. The operation of these systems helps execute timely interventions to prevent accidents, improve emergency response times, and inspire a safer urban transportation ecosystem.

We have seen that AI-based traffic management systems play a pivotal role in ensuring sustainability in resource allocation. They also promote equitable access to transportation infrastructure. They accommodate diverse transportation modes, including public transit, cycling, and walking, alongside traditional vehicular traffic. This ensures that underrepresented and underserved communities, who are often reliant on public transit, benefit from strengthened traffic mobility. AI-boosted traffic systems support sustainability by minimizing carbon emissions by reducing idle time and fuel consumption through adaptive signal control. They integrate multimodal transport systems, thus facilitating the prioritization of environmentally friendly options such as electric vehicles (EVs), car-sharing, and non-motorized transit like bicycles. Therefore, they contribute to a more verdant urban ecosystem.

*Historical Context and the Evolution of AI in Transportation*

The seeds of AI in transportation were sown in the mid-20th century, primarily in the realm of automation. One of the main manifestations of this was computerization of the traffic signal. The first steps toward automating traffic control were taken in the early 20th century with the initiation of electromechanical systems. These systems use a combination of mechanical timers and electrical relays to control traffic signals. A breakthrough came with the development of the traffic signal controller. This device was equipped with a microprocessor. It analyzes traffic data from various sensors, such as vehicle detectors and pedestrian push buttons. We are currently moving towards Intelligent Transportation Systems (ITSs) and multi-modal systems. Smart traffic lights and AI-driven signal control systems are an intricate part of ITSs. Vehicle-to-infrastructure (V2I) and vehicle-to-vehicle (V2V) [20] communications, as parts of smart signaling, are also implemented within ITS and help in seamless data exchange between connected vehicles and infrastructure elements. ITSs also involve integration of distributed, edge & cloud computing for real-time traffic management in high-traffic environments. ITSs, in various stages of development, have been implemented in global cities like London and Shanghai [21]. An ITS forms a part of the broader Smart Cities initiative, key to which is the Internet of Things (IoT) technology.

Another manifestation of this was the automation of highway toll systems. The initiation of computerized highway toll systems dates back to the early 1970s when electronic toll collection (ETC) prototype experiments were carried out. By the late 1980s, ETC systems gained traction in Norway. Throughout the 1990s, the use of ETC systems expanded globally. Technologies such as Radio Frequency Identification (RFID), transponders, electronic toll collection lanes, and centralized systems were crucial in automating toll collection processes. The institutive process involved pilot programs, infrastructure development, public education campaigns, and system integration to ensure the smooth adoption of ETC systems. These computerized toll systems brought about various benefits, including reduced chance of traffic jams, finessed productivity, amplified safety, and enhanced convenience for drivers. In recent years, advancements in highway toll system technology have further developed toll collection methods. Contactless payment integration through mobile wallets and NFC tags, multi-lane free flow systems with advanced sensors, dynamic pricing strategies, integration with ITS, and electric vehicle charging station integration have been implemented.

One of the most widely studied topics in the realm of transport automation is AVs. The history of AVs dates to 1939, when General Motors unveiled an exhibit called Futurama in New York. It was a diorama installation that featured automated highways. From 1980 to 2003, university research centers, often in collaboration with transportation agencies and automotive companies, conducted studies on autonomous transportation. The NavLab project, set up at Carnegie Mellon University around the mid-1980's, involved the development and testing of AVs, starting from NavLab 1 up to NavLab 11 [22] [23]. DARPA's Grand Challenges boosted AV technology. Stanley, an innovative robot vehicle, as a part of these challenges, won an AV race in 2005 [24]. Google's Driverless Car initiative made significant advances in commercializing AV technology starting the early 2010s, and the initiative continues to grow.

The 1990's saw the beginning of the age of ITSs. ITSs integrated sensors, communication networks, and AI algorithms within the realm of transportation. They utilized methodologies like Adaptive Traffic Control

Systems [25] and Advanced Driver Assistance Systems [26]. The former dynamically used traffic data to adjust signal timings. The latter involved technologies like cruise control, lane departure warning, and emergency braking. ITSs are important components of modern urban planning, as they serve to reorganize traffic flow for the better, magnify road safety for people and vehicles, reduce costs and promote economic growth through amplified efficacy and precision and efficacious resource allocation and utilization, and bring in sustainability, holisticness, equity, inclusivity, affordability and accessibility for all.

Currently, commercially viable technologies that enable vehicles to perceive their surroundings, make decisions, and act without human input, are being researched. Challenges like safety, liability, and privacy, the need for infrastructure compatibility, and overcoming public skepticism against reliance on technology, are being tackled head on. Innovations like connected vehicles, personalized mobility services, and sustainable transportation, keys to a more intelligent, economical, safe, and eco-friendly traffic paradigm, are the guidelines for future research.

## Fundamentals of Traffic Systems

Traffic systems refer to the aggregate of infrastructure, rules, and technologies that govern the movement of vehicles and pedestrians on roadways. Traffic systems, as a topic, broadly subsumes roads and highways, traffic signals and signs, transportation management systems, and traffic laws, regulations and security. Traffic systems, while essentially a general term, covers only road traffic within its ambit, and not air traffic and railways. Air traffic is covered under aviation (aviation engineering), while railway traffic is covered under rail transportation (rail transportation engineering or rail traffic engineering). Traffic systems are founded upon the principles of traffic flow theory and modeling, concepts which provide a framework for understanding the behavior of vehicles and drivers on roadways.

*Overview of Traffic Flow Theory*

Traffic flow theories are mathematical models that seek to explain how vehicles, drivers, and infrastructure interact with each other's presence on the road. They are used by transportation engineers to stimulate (model), understand and analyze traffic flow (Figure 2). They help in finding ways to introduce and redesign infrastructure elements like signage, traffic lights, traffic cameras and markings.

The scientific study of traffic flow can be traced back to the early 20th century. As urbanization and industrialization led to a rapidly growing automobile ownership and thereby the frequency of traffic jams, there was a growing need to understand and manage traffic flows precisely. One of the earliest significant contributions to the field came from Bruce Greenshields in the 1930s [27].

His work involved applying probability theory to understand road traffic flow and it laid the foundation for understanding how individual vehicles interact and influence traffic flow. Since then, the field has continued to evolve, with researchers developing more complex models, collecting and analyzing vast amounts of traffic data, and applying advanced technologies like AI (including machine learning) to upgrade the current pattern of traffic management.

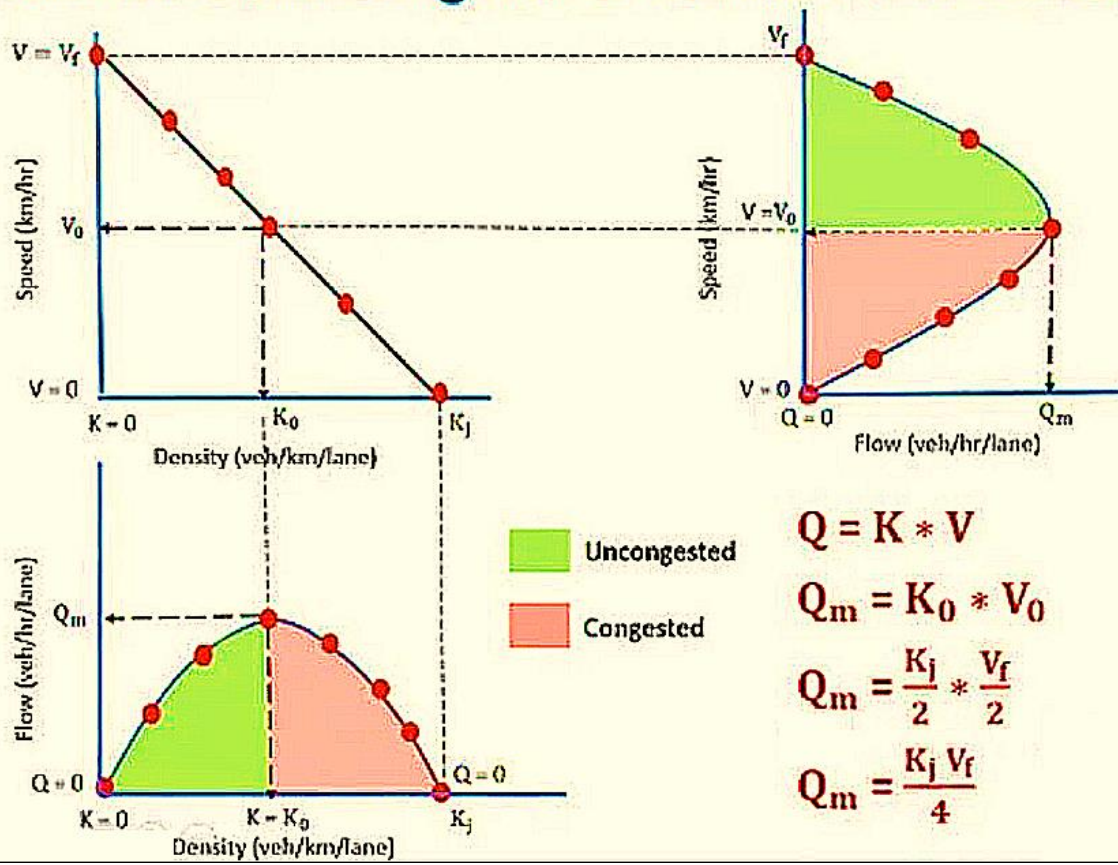

Figure 2: Diagrammatic Representation of Traffic Flow Theory. Here, Q stands for Traffic Flow, K stands for Traffic Density and V stands for Traffic Volume. $Q_m$ represents the maximum traffic flow. Traffic flow struggles when traffic density and traffic volumes are higher than the optimal levels or the threshold levels ($K_0$ and $V_0$, respectively). Traffic speeds are also dependent on K and V, with the traffic achieving an optimal speed when the traffic volume is $V_0$, and the traffic density is $K_0$. (Borrowed/Adapted from Priya Rai) [28].

*Traffic Management Strategies*

Traffic management strategies are approaches and techniques used to coordinate vehicular flow, minimize collisions and the likelihood of roads shutting down and cause furtherance in the safety of motorway networks. This falls under traffic engineering, transportation engineering or highway engineering. Some of the key traffic management strategies are discussed here.

Traffic Signal Control
Signal control is a fundamental strategy in traffic management. It uses traffic lights and signaling devices to regulate vehicle flow at intersections. It plays a key role in urban transportation by activating a better traffic flow, lowering the probability of traffic jams, and stirring road safety levels. The primary objectives of signal control include maximizing traffic throughput, minimizing delays, boosting safety, and mitigating road closures. At busy city intersections, traffic lights are synchronized to allow a maximum number of vehicles to pass through during a green light. This helps reduce the time vehicles spend idling.

Fixed-time signal control is a traditional method of signal control. In it, signals operate on a predetermined cycle [29]. This fact makes fixed-time signal control suitable for intersections with relatively stable traffic patterns, such as in residential neighborhoods. Nonetheless, this method can be uneconomical during peak hours when traffic is highly variable. In contrast, actuated control uses sensors to detect vehicles and adjust signals in response to actual traffic conditions. This approach offers greater flexibility, making it ideal for areas where traffic flow changes frequently, such as near commercial centers. Intersections near commercial areas like shopping malls often use actuated signals to prevent unnecessary delays when traffic is light.

A more advanced approach is adaptive signal control. Adaptive control dynamically adjusts signal timings based on real-time traffic data [30]. This system is particularly useful in large cities where traffic patterns fluctuate throughout the day. Adaptive signals in metropolitan areas respond to rush-hour surges by adjusting light cycles to ease traffic jams in real time. Though highly lucrative and beneficial, adaptive systems require advanced technology and data collection to operate functionally and as intended. Traffic signal priority is a signal control technique which accords precedence to specific vehicles like emergency services and public transit. For instance, an ambulance may trigger signal changes to get through traffic faster. This reduces emergency response times in critical situations.

Synchronized traffic signals along major corridors create a green wave effect. The green wave effect allows vehicles to move smoothly through a series of lights with minimal stops. This approach is implemented on major urban roads where traffic is heavy. Coordinating the signals causes a burgeoning in the plausibility of the movement of road vehicles becoming much more cost-effective.

The accuracy and efficacy of signal control depends on such factors as traffic volume, composition, intersection geometry, and pedestrian traffic. A crowded downtown intersection with high pedestrian activity and multiple turning lanes begs a more sophisticated signal control strategy than a simple suburban intersection with light traffic and fewer turning lanes. As urban areas expand and traffic patterns become more complex, signal control systems should also adapt. The integration of signal control with ITSs and connected vehicles represents a future trend that targets to make traffic organization better [31]. The use of big data and analytics is becoming essential for rationalizing signal control strategies. Traffic data collected from various sources informs better decision-making. It ensures that signals are adjusted based on real-time conditions.

Congestion Pricing

Congestion pricing is a traffic management strategy that charges drivers a fee for using roads during peak hours or in highly jampacked traffic zones. The primary goal of enforcing congestion pricing is to reduce traffic gridlocks through discouragement of driving during busy times. This system typically relies on electronic tolling. In this, vehicles are automatically charged as they pass through designated areas. Dynamic pricing is another feature which is a part of some congestion pricing systems [32]. Under this, fees are adjusted based on the level of vehicular queuing.

Congestion pricing incentivizes people to find alternative routes, travel at off-peak hours in lieu of rush hour, and use public transportation. This helps cut back the number of vehicles on the road and rein in emissions to liven up sustainability and eco-friendliness. Congestion pricing generates revenue for cities and transportation agencies. This revenue is usually reinvested in infrastructure developments and in boosting public transportation systems, such as funding new bus routes. This expands and further encourages public transit options for commuters.

Nevertheless, congestion pricing comes with its challenges. Concerns related to social equity arise in relation to congestion pricing since low-income drivers are disproportionately affected by it. Such drivers have fewer alternatives available to them. This makes the fees feel unfair, especially if public transportation options are limited in such areas. The costs of implementing a congestion pricing system are also a consideration. Putting such initiatives in place requires investments in technology and infrastructure such as electronic toll systems, traffic monitoring tools, RFID technology, and enforcement mechanisms [33]. Public acceptance of programs like this is commonly a barrier [34]. Many drivers resist the idea of paying additional fees for roads they are accustomed to using without charge. A number of city administrations have faced pushback and backlash from drivers unwilling to participate in such endeavors.

There are successful examples of congestion pricing in certain major cities. A Congestion Charge rule has been in implementation in London, United Kingdom. In this, drivers are charged a fee to enter a certain zone called the congestion charge zone, during peak hours. Singapore has a system of Electronic Road Pricing (ERP) [35]. Under this, drivers are charged when they want to enter certain specific areas during peak times. This system uses electronic tolls. Both systems have demonstrated the serviceability, supportability, and utility of congestion pricing in reducing traffic and encouraging alternative forms of transportation. They also help generate revenue for urban development projects in their respective areas.

Tolling
Tolling is another traffic management strategy where drivers are charged a fee to use certain roads, bridges, and tunnels. The main goal is to manage traffic flow, generate revenue for infrastructure maintenance and advancements, and, in some cases, discourage the overuse of frequently clogged traffic routes. Toll roads are common in many countries. In the United States, the New Jersey Turnpike [36] and Florida's Turnpike [37] are examples of toll roads that use fees to maintain infrastructure and manage traffic. In Europe, countries like France and Italy use toll systems extensively on their major highways.

Tolling can be implemented in several ways, including traditional toll booths, where drivers pay manually, and modern electronic tolling systems that automatically charge vehicles. Variable tolling, also termed congestion tolling or dynamic pricing, is a way of tolling in which fees upsurge during peak hours in highly areas prone to traffic gridlocks.

One of the key benefits of tolling is its ability to generate revenue for transportation projects. The funds collected can be used to maintain roads, develop infrastructure, and expand public transportation systems. Tolling also helps decrease roadblocks by discouraging excessive use of certain routes. When drivers must pay to use busy highways, a preference for alternative transportation routes and modes automatically grows.

Tolling can take various forms, depending on the goals of the transportation system. Congestion-based tolling involves charging higher fees during peak traffic times to mitigate traffic blockages. For example, tolls on the I-66 express lanes in Northern Virginia fluctuate based on the level of traffic, intensifying during rush hour to deter drivers and to cause a decline in vehicle miles traveled (VMT) [38]. Distance-based tolling charges drivers based on the distance traveled on a toll road. This ensures that long-distance travelers pay more for using the infrastructure more. Flat-rate tolling, on the contrary, charges a fixed fee for road usage, regardless of time and distance of usage.

All the lanes on I-66 Beltway have shifted from high-occupancy vehicle (HOV) lanes to high-occupancy toll (HOT). HOV lanes are lanes restrictions on use of single-occupancy vehicles (SOV) apply in order to endorse ridesharing. HOV lanes have been set up at various places across US and the rules that govern their use vary [38]. Vehicles carrying two or more people are generally permitted on HOV lanes. At times, motorcycles and vehicles that use alternative sources of energy are also permitted, thereby promoting sustainable transportation. Curbs on the use of HOV lanes can operate either round-the-clock or just during rush hours and known periods of high traffic logjam possibility. Where there is excess capacity on HOV lanes, HOT lanes may be implemented, where motor vehicles carrying only one person can use these lanes for a fee. In regions where these lanes serve as major arteries, the use of such lanes by proscribed vehicles may be allowed for a fee, thereby implementing congestion pricing through tolling.

## AI Technologies and Their Applications in Traffic Management

AI technologies which involve statistical analysis, machine learning, deep learning, reinforcement learning, unsupervised learning, semi-supervised learning and so on, are gradually becoming the new normal in traffic management. With accumulating human population and an escalation in traffic pressure on urban areas, AI-powered technologies are providing the necessary innovation to cohere traffic flow in a way that suits the demands of metropolitan zones (Figure 3). These innovative technologies upgrade transportation supportability, value, and resilience for road users.

Road traffic prediction implies utilizing traffic data dynamically to forecast the traffic situation in a particular region given various circumstances like an accident, an unusual or adverse weather event, rush hours, the start of road construction work or any other roadside infrastructure development project and so on. Precise and well-timed forecasts provide relevant support to commuters, drivers, pedestrians, cyclists, etc. along with traffic management administration, and to traffic policy advisors and transport infrastructure advisory panels. These forecasts serve to assist transportation operations and traffic route scheduling. They also inform and serve as guides to traffic infrastructure expansion.

This prediction can be based on the output of existing mathematical models of traffic which have been constructed taking into account various parameters that can influence traffic like the time of the day, the occurrence of an unexpected event and so on. They are simulation models. This prediction can also be made based on the algorithmic analysis of real-time data, for instance, the number of cars and so on. The data-driven approach, based on machine learning algorithms, deep learning algorithms and other intelligent methods, has been found to be more reliable since the model-based approach is based on static models. However, a hybrid approach has been proposed as the best [39].

In this way, machine learning algorithms, such as neural networks, regression models, and deep learning, are becoming central to predicting traffic flow. These algorithms can be designed to develop the capability to analyze vast datasets, including historical traffic patterns, weather conditions, and real-time traffic data, to forecast gridlocks and suggest optimal routes. Deep learning models predict how traffic will change throughout the day. This allows for better traffic signal timing and route planning. Real-time traffic monitoring systems, supported by machine learning, dynamically respond to sudden changes, such as accidents or road closures. This ensures a more resilient traffic management.

AI-enabled computer vision plays a vital role in traffic management. This is especially in the case of AVs. Computer vision systems like CNNs and vision transformers (ViTs) detect, identify, and track vehicles in real-time. This helps traffic management systems and ITSs gauge vehicle movements and traffic obstruction levels. In AVs, computer vision, with the use of cameras, LiDARs and sometimes other sensors like infrared sensors and radars, enables cars to "observe" their surroundings, giving the cars positional awareness. The cars can recognize and detect objects, other vehicles, pedestrians, road signs, traffic signals, guardrails, median dividers, lane markings, and other features of the road, and adjust their own driving accordingly. When an obstruction, a red or yellow traffic signal, a pedestrian, a slowing vehicle in the same lane as the AV, and so on, is detected on the road, the vision system detects it and passes on the information to the control unit, which changes the velocity of the car accordingly. Sometimes AI systems are integrated with AVs to provide actionable real-time data on traffic conditions and enable a more relaxed navigation through crowded urban environments. The images of the surrounding environment captured by the sensors of the

AV are segmented to give the control unit in the AV a thorough grasp on its vicinity. This extenuates the likelihood of accidents.

Reinforcement learning (RL) helps in implementing various smart tasks (AI tasks) in traffic management. Adaptive traffic signal control is one such task where RL can prove beneficial, as these systems can be precisely and incrementally strengthened via trial-and-error and reward-and-punishment methodologies [40]. Adaptive traffic signal control systems are those traffic signal control systems which have been able to learn how to adjust and rationalize traffic signal timings based on real-time conditions. RL algorithms make decisions by continuously interacting with the traffic environment. They repeatedly adjust signal timings to ameliorate vehicle flow. One case study from Pittsburgh [41] exemplifies how adaptive signal control systems have successfully managed complex urban traffic patterns.

In traffic management NLP is used in dynamic messaging signs and communication with drivers [42]. NLP algorithms analyze and generate human-readable messages on digital road signs to alert drivers about traffic conditions, accidents, or route changes. The use of real-time data ensures that the messages displayed are timely and relevant. This facilitates communication between traffic management systems and drivers. NLP is integrated into voice-activated navigation systems such as Google Maps. This provides drivers with traffic updates and alternative route suggestions in natural language.

AI technologies are also instrumental in incident detection and management. Using data from traffic cameras, sensors, and GPS, AI systems predict and detect accidents and other incidents in real time. Machine learning models analyze patterns and identify potential high-risk areas. This empowers authorities to take necessary preventive measures. In the event of an accident, AI systems automatically coordinate emergency response services to enable faster response times. AI-operated platforms in cities like Singapore detect traffic incidents and send automatic alerts to emergency services [43]. This resolves emergency response times and mitigates the influence of traffic disruptions on emergency response.

Edge and cloud computing utilize distributed computing and are being applied to traffic and transportation management. These technologies empower the proficient collection, processing, and analysis of vast amounts of traffic data from various sources. Decisions can be made quickly in high-traffic environments if AI models are deployed at the edge. In this way, edge computing maximizes throughput and minimizes latency for timely interventions. It helps in dynamic traffic monitoring and real-time decision-making.

Public transportation systems also benefit from AI-driven optimization. AI technologies are being used to smoothen bus and rail scheduling. These technologies ensure that public transport services run on time and adapt to changing conditions. Intelligent route planning systems provide commuters with real-time updates on the best routes and travel options. This makes public transportation more reliable and appealing. This in turn contributes to proliferation of ridership and dropped reliance on private vehicles [44] [45] [46].

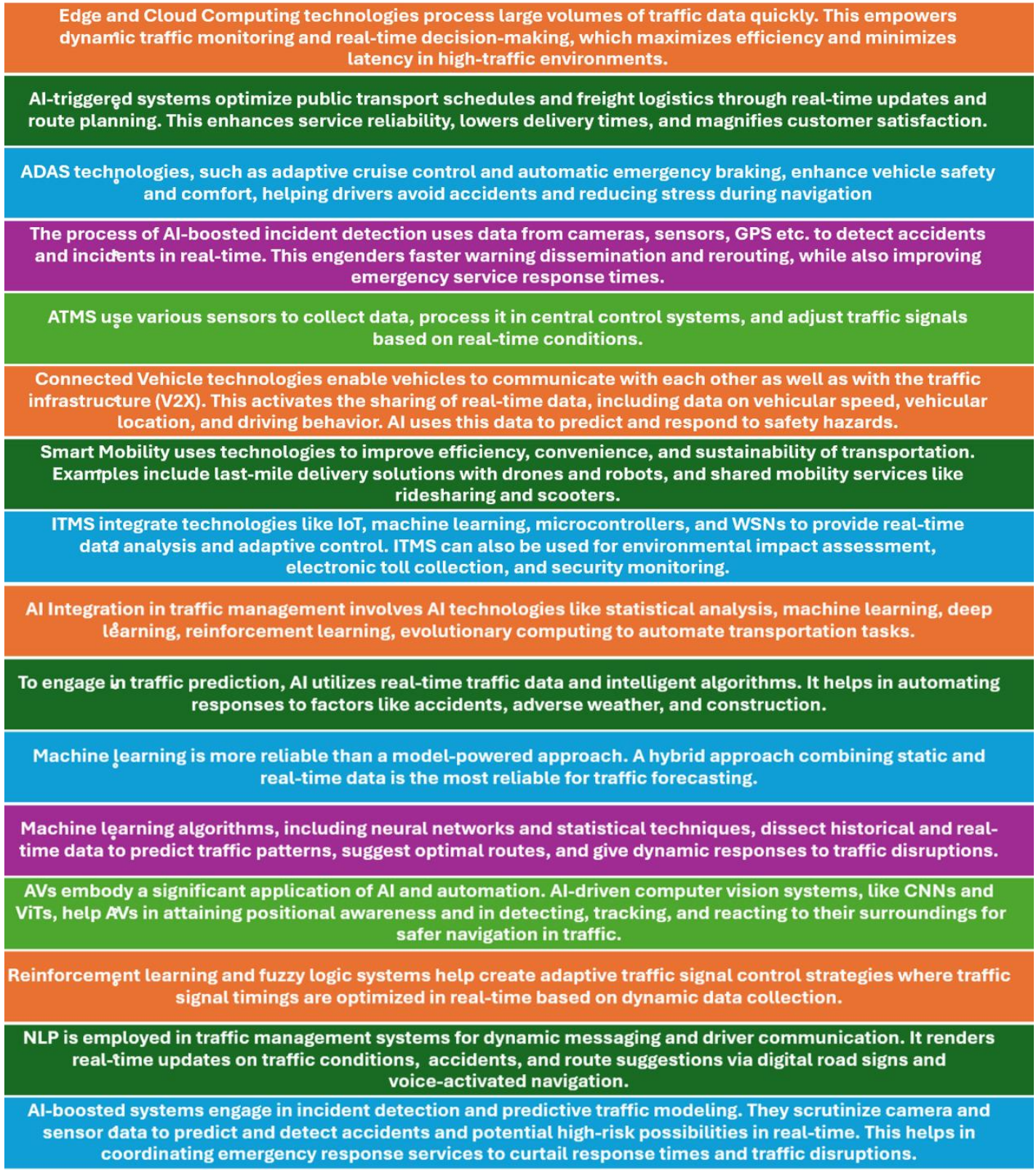

Figure 3: A summary of applications of AI in traffic management

AI is contributory in optimizing the movement of goods in the realm of freight and logistics. AI supports truck platooning, where a group of trucks travels in close formation, reducing fuel consumption and extending safety. AI-fired logistics planning helps companies fine-tune delivery routes and schedules. This ensures timely, judicious and cost-efficient transportation of goods.

*Advanced Transportation Management Systems*

Advanced Transportation Management Systems (ATMS) are critical components of extant urban infrastructure. They work in tandem with Smart Cities. They enable better traffic management and flow, moderated possibility of traffic snarls and a strengthened productivity of transportation networks.

ATMS engage various traffic sensors to collect essential data so that they can undertake data-driven decision-making. Inductive loops, embedded in roads, are sensors which detect vehicles by measuring changes in electrical current. Magnetic sensors can also be employed within ATMS to detect vehicles. Magnetic sensors record changes in the earth's magnetic field caused by passing cars. Cameras provide visual data like vehicle counts and speeds. A LiDAR uses laser beams to create 3D maps of the environment. Radar systems, using radio wave reflections, are equipped to detect the presence of vehicles under various unusual weather conditions. To be analyzed, all the data collected through these sensors is carried to the hub, that is, the central control system. Dedicated Short-Range Communications (DSRC) play a vital role in V2V and V2I communications [47]. WLAN (Wireless Local Area Network), 4G and 5G Networks and fiber optic networks ensure high-speed, reliable data transmission between field devices and central systems.

Central control systems, such as Transportation Management Centers (TMCs), serve as the heart of ATMS. Real-time traffic data is processed and analyzed in these hubs. Traffic control software uses sensor data to monitor traffic flow, detect incidents, and adjust signal timings. Data analytics tools within these systems analyze large volumes of information to detect patterns. Traffic control devices such as traffic signals, variable message signs (VMS), and ramp meters regulate vehicle flow and provide drivers with real-time information. Dynamic speed limits adjust according to traffic conditions [48]. Traffic barriers physically manage lanes during emergencies and construction activities.

AI enables predictive traffic modeling, incident detection, and the finetuning of signal timings based on real-time information. The IoT connects traffic sensors, vehicles, and infrastructure to allow real-time data sharing. IoT technology, therefore, makes the traffic management systems more responsive to traffic conditions. More vehicular processes are being automated so that cars can take decisions independently of drivers using sensor-collected data, especially when manual decision making can be slower and subjective and may lead to endangerment of lives. Vehicles that are becoming cumulatively more autonomous are rapidly turning into an essential part of ATMS. They interact with traffic systems in a way that suffices to crucially amplify traffic flow and road safety.

Floating Car Data (FCD) is a term applied to the collection of real-time vehicular data, like car location, speed and direction of travel. This data is collected throughout the road network by geolocating the vehicle using Global Positioning System (GPS)-enabled devices like smartphones present in the vehicle. After this data has been mined, it is sent for processing. Its analysis reveals handy practical findings like the status of traffic and is used to derive insights like alternative routes in case of potential disruptions and bottlenecks. This information is transmitted to the drivers so that they can be warned and be allowed to take the most prudent decisions based on the knowledge gained.

Big data analytics tools process the enormous amounts of data collected. Identification of traffic trends and anomalies helps make better informed decisions. Cloud computing provides scalable storage for the AI models, for the large amount of data that is continuously generated in these systems, and, most importantly, for the results from processing the data. They also allow parallel processing and the sharing of superior processing power across systems. This enhances processing speeds, strengthens the flexibility of traffic management systems, and supports real-time decision-making. ATMS render a smarter and more adaptable approach to handling the complexities of urban traffic.

*Advanced Driver Assistance Systems*

Advanced Driver Assistance Systems (ADAS) are a suite of technologies that are a part of ITS. They are designed to expand vehicle safety and comfort by assisting drivers with various tasks. These systems rely on sensors, cameras, and control units to monitor the driving environment. ADAS intervene whenever necessary to prevent accidents [44]. Radar systems offer information about distance, speed, and direction of nearby objects as well as of the system they are a part of. Cameras capture visual data of the road and surrounding areas and are another cog in the mechanism of positional awareness apart from serving as an additional eye for the driver. LiDARs provide precise distance and position details. Ultrasonic sensors utilize sound waves to detect objects. The control units process data from these sensors and determine appropriate actions, which are then carried out by actuators that manage functions such as braking and steering.

ADAS encompasses a wide range of functionalities. Adaptive Cruise Control (ACC) helps a driver maintain an appropriate distance from vehicles ahead using sensors. It adjusts the vehicle's speed automatically. Lane Departure Warning (LDW) alerts drivers when the vehicle drifts out of its lane, while Lane Keep Assist (LKA) system applies corrective steering and centralizes the vehicle's position within the lane. A frontal collision warning (FCW) system is a driver assistance technology that alerts the driver when their vehicle is approaching a collision with an on-road object such as another vehicle. Automatic Emergency Braking (AEB) intervenes when a collision seems imminent. It controls the brakes and applies them if the driver does not react in time. Rear Cross-Traffic Alert (RCTA) detects oncoming vehicles during reverse maneuvers. Blind Spot Monitoring (BSM) alerts the driver to the vehicles that are in the driver's blind spots. Hence, BSM and RCTA reinforce people's awareness during parking, reversing and changing lanes.

Additional features like Head-Up Display (HUD) project important information onto the windshield. Night Vision raises visibility in low-light conditions. Traffic Sign Recognition (TSR) uses cameras and AI software to identify and display traffic signs to the driver. Traffic Sign Recognition identifies road signs and displays and simplifies them for the drivers, especially the drivers who face some ophthalmic or optical trouble or other problems like dyslexia. Driver Drowsiness Monitoring detects signs of fatigue and prompts alerts to ensure benign driving. Driver attention monitors keep a watch on the steering wheel inputs and the driver's face. Adaptive Cruise Control is also an ADAS technology.

The benefits of ADAS are substantial, with an uptick in traffic safety being the most critical. These systems assist drivers in avoiding hazards and responding more swiftly to potential threats. ADAS also supplements the general driving experience. It reduces drivers' stress by taking over several driving-related responsibilities using capabilities such as adaptive cruise control. These technologies contribute to a surging fuel economization. As ADAS continues to evolve, more advanced features and capabilities, such as digital twinning, for a more nuanced understanding of the environment by the smart processing setup through real-time lifelike representation of objects in the vicinity of the vehicle in [49], are regularly being introduced.

Google Maps is a widely used navigation app. It is a service by Google which utilizes GPS technology to determine a user's location. GPS involves a network of satellites orbiting the Earth. They send real-time signals that can be received by GPS-enabled personal devices like smartphones. These devices calculate their position and velocity based on the time difference between the signals received from different satellites. Google Maps then combines this location data with a vast database of roads, landmarks, and traffic information to provide accurate maps and directions. The app also uses algorithms to analyze real-time traffic data. This allows it to suggest the most prudent routes and provide timely updates on traffic conditions.

Google Maps offers a valuable feature: voice-enabled directions. This technology utilizes AI features such as natural language generation (NLG) and speech generation capabilities. It also deploys the usage of GPS to provide drivers with real-time guidance. As users steer through road networks, Google Maps tracks their location using the GPS signal. This information is then used to calculate the optimal route based on factors like traffic conditions, road closures, and construction zones. The app then provides clear, step-by-step directions, guiding the drivers through the journey. These directions are given through an automated voice to let drivers concentrate their sights on the road. If a traffic jam is detected ahead, Google Maps suggests alternative routes, advises as to which one is the best, and provides voice instructions to traverse it. This feature enhances the driving experience, slashes stress levels, and helps ensure secure and sensible travel.

*Connected Vehicle Technologies*

AI-based traffic management systems involve connected vehicle technologies as a cornerstone in inducing trustworthiness and pushing down the necessity for exercising caution within urban transportation networks. These systems amalgamate V2V, V2I, V2N (vehicle to network), and V2P (vehicle to pedestrian) (combinedly called V2X) communication protocols [50][51]. V2N technology enables vehicles to communicate with the internet and with other entities that are connected to the internet using wireless and 4G/5G networks. It is V2N communications that enable IoT interactions and access to various kinds of data and information for the vehicles. They also enable vehicles to transmit signals over distances, for example, to central transportation hubs.

V2P communication allows vehicles and pedestrians to directly pass signals to one another. This approach incorporates an extensive collection of the classes of pedestrians including people walking or jogging, toddlers being pushed in prams or strollers, wheelchair users and other para-passengers and para-pedestrians relying on personal mobility devices, passengers boarding or deboarding from public transport including buses, trains, PRT, trams, or infrastructure associated with public transport like metro train stations, subway stations, ferry ports, cable cars, paratransit etc., and bicycle riders. Vehicle-to-satellite (V2S) technology enables satellite communication for vehicles, which is an alternative to WLAN and 4G/5G communication.

V2I communication enables real-time data exchange between vehicles and infrastructure elements such as traffic signals, road signs, and central control systems. V2X communication systems give the transportation paraphernalia the capacity to harness the data generated by vehicles which are a part of these systems. The data includes information on speed, location, and driving behavior. Using such data, AI algorithms predict and respond to potential safety hazards proactively [45]. In scenarios where a connected vehicle detects sudden braking, this data is instantaneously relayed to the traffic management system. The response can be implemented in the form of adaptive responses such as adjusting traffic signal timing and alerting nearby vehicles to take precautionary measures. This creates a more responsive and predictive approach to safety management.

*Smart Mobility*

Smart mobility encompasses a diverse range of technologies and services that aim to make transportation more efficient, convenient, and sustainable for users. It extends beyond traditional modes of transportation by integrating advanced digital solutions to adapt to urban mobility and logistics. This ecosystem includes not only the expedient movement of people but also the smart delivery of goods, innovative supply chain management, and the use of technology to harmonize booking, payments, and shared mobility services such as ridesharing, scooters, and bicycles.

One key aspect of smart mobility is the evolution of last-mile delivery solutions. E-commerce continues to grow. In this light, there is a multiplying demand for timely and expedient delivery of goods. Smart delivery solutions, such as autonomous delivery robots and drones, have been tested and implemented to curtail traffic blocks and pollution associated with traditional delivery methods. These technologies are designed to maneuver urban environments autonomously. They deliver packages directly to customers' doorsteps. AI-powered route modification and fine-tuning and real-time tracking systems enable logistics companies to manage deliveries more judiciously. This diminishes delivery times and costs while raising customer satisfaction.

In the supply chain sector, smart mobility solutions are leading to evolution in the way goods are transported and managed. IoT sensors and blockchain technology are being used to track the movement of goods in real-time. This enables transparency and flourishing productivity throughout the supply chain. This level of visibility helps companies anticipate potential disruptions, reorganize and automate inventory management, and shrink wastage. Smart logistics platforms that integrate machine learning and soft computing tools, are also being used to predict demand, systematize warehouse operations, and plan for dynamic routing. This makes the entire supply chain more agile and responsive to changing market conditions.

The advent of shared mobility services, such as ridesharing and shared micromobility options like scooters and bicycles, has been witnessed in extant civic scenarios. This has revolutionized urban transportation. These services provide flexible, on-demand transportation options that reduce the need for private car ownership, alleviate traffic blockages, and result in improvements in ecological sustainability. Integrated platforms that combine booking and payment for various modes of transportation – whether cars, bikes, or scooters – are becoming popular. They allow users to plan and pay for their entire journey in a seamless manner and encourage a shift towards more sustainable and multimodal transportation habits. These services accompany real-time data analytics and user feedback systems. This helps operators augment fleet distribution and availability based on demand patterns.

Payments and booking systems are also an integral part of the smart mobility framework. Digital payment solutions, inter alia mobile wallets and contactless payments are becoming standard in public transit systems. This enables faster, facilitated, and more secure transactions. Integrated mobility-as-a-service (MaaS) platforms enable users to plan, book, and pay for multi-modal trips through a single application. This simplifies the user experience. This also promotes the use of public and shared transportation options over private vehicles.

*Intelligent Traffic Management Systems*

The development of ITMS heralds a new era in traffic management. ITMS are a form of ATMS. These systems integrate technologies like IoT, machine learning, microcontrollers, wireless sensor networks (WSNs), and fuzzy logic. These technologies are being integrated into smart infrastructure to provide real-time data analysis and adaptive control [43] [52].

IoT enables interconnected devices to collect and transmit data from various traffic-related sources, such as traffic signals, cameras, and vehicle sensors. The decisions taken at the central hub are implemented at the edge through IoT. IoT supports environmental monitoring by tracking air quality and noise pollution. Microcontrollers and WSNs form a part of IoT. They are essential for the deployment of smart traffic solutions at a granular level. Microcontrollers are used to process data locally. They also manage devices such as traffic lights and sensors. They are key to enabling the execution of real-time commands based on data inputs. WSNs facilitate communication between different components of the traffic management system and ensure coordination.

Machine learning algorithms analyze vast amounts of traffic data, detect anomalies, and predict future traffic conditions. Machine learning forms the basis of autonomous traffic control systems that operate with minimal human intervention. Fuzzy logic deals with the uncertainties and variability inherent in traffic systems. It handles imprecise and constantly changing information. It makes decisions in complex, real-world situations where clear-cut rules are not applicable. Fuzzy logic systems can make complex choices and competent decisions based on various degrees of traffic density and flow (Figure 4).

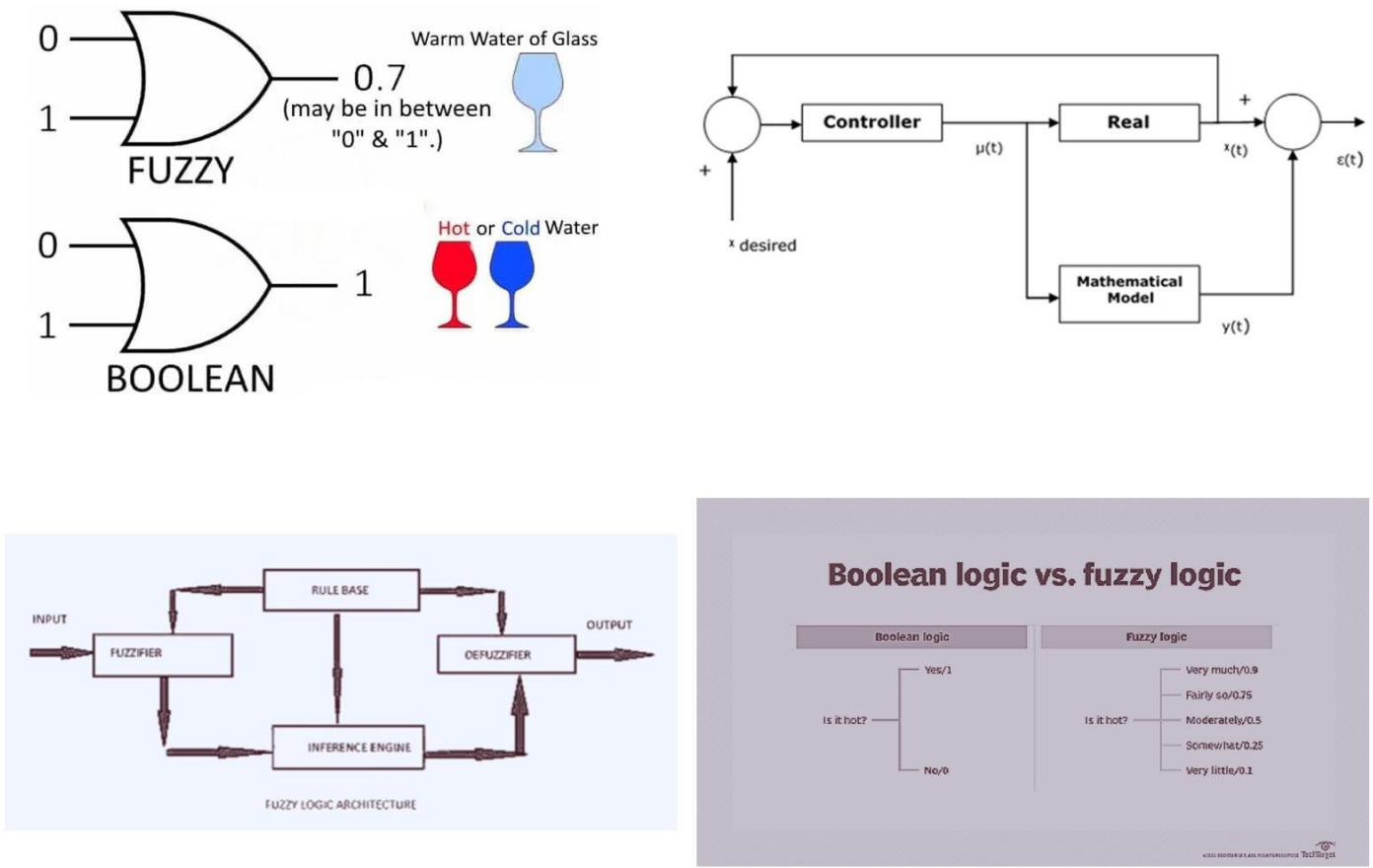

Figure 4: Basic Description of Fuzzy Logic (Borrowed/Adapted directly from the Internet)

The applications of ITMS extend beyond traffic control. ITMS is advantageous in areas such as environmental impact assessment, electronic toll collection, and security monitoring. Environmental impact assessment involves using data from the ITMS to analyze the effects of traffic on air quality and noise levels. This provides insights that enable sustainable urban planning. Electronic toll collection systems use automated processes to reduce bottlenecks at toll points. Anomaly detection and illegal activity identification are crucial for maintaining security. Smart systems like ITMS have the capacity to identify suspicious behavior, such as illegal parking, signal jumping and traffic violations. ITMS can also support security monitoring by integrating with surveillance systems to ensure road safety and compliance with traffic laws.

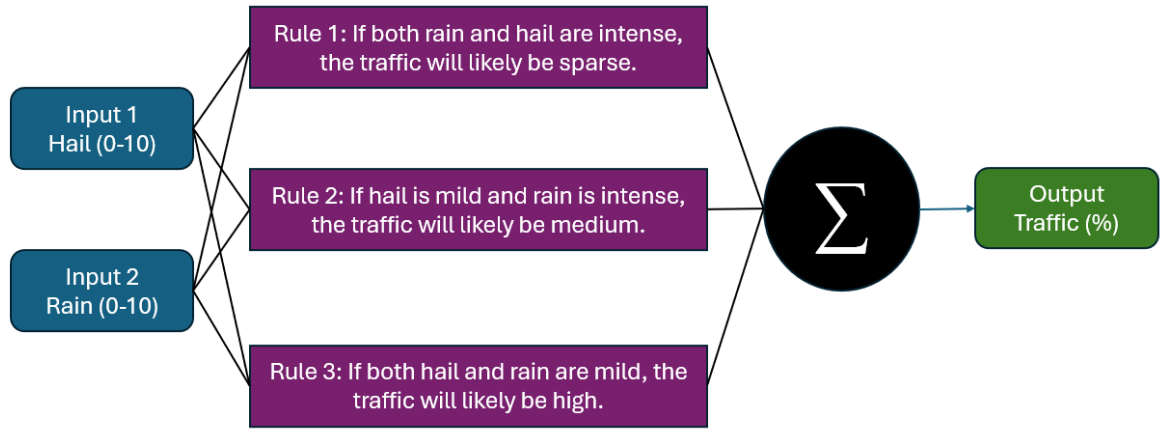

**Figure 5: An example fuzzy system for traffic prediction based on weather conditions (given that there exists a definitive hailstorm). It consists of two inputs, three rules, and one output. The inputs are crisp (non-fuzzy) 10-point Likert scale quantities; the rules are fuzzy and evaluated in parallel using fuzzy reasoning; the results of applying the rules on the inputs are combined and defuzzified; and the output is a non-fuzzy percentage.**

*Additional Applications*

The application of AI in transportation is heavily supported by various machine learning techniques. Supervised learning models such as support vector machines (SVMs), probabilistic neural networks (PNNs), graph neural networks (GNNs) and decision trees are used for prediction and classification tasks in transportation management. These models rely on historical data collected from sensors to forecast short-term traffic patterns. Unsupervised machine learning methods such as clustering are employed for more complex problem-solving. Deep learning techniques, which include artificial neural networks (ANNs), have been applied in predicting traffic conditions and assisting in decision-making that would aid with road infrastructure planning, incident detection, and so on [53].

Optimization algorithms including genetic algorithms (GAs) and simulated annealing (shortened to SA, it is a probabilistic algorithm originally designed to solve the Traveling Salesman Problem; preferred for its utility in estimating the global optimum of a function [54] aid in solving issues arising in the design and management of transportation networks. These algorithms are inspired by biological processes. They efficiently imbibe, navigate, and analyze the complexities of urban design. They offer serviceable solutions to the challenges posed by congestion and infrastructure limitations.

The development of swarm intelligence systems such as Ant Colony Optimization (ACO) and Bee Colony Optimization (BCO) draw inspiration from the collective behavior of ants and bees to solve optimization problems related to routing and traffic flow management. In urban planning, complexity of infrastructure development requires optimization models like ACO and BCO, which have demonstrated capability in providing solutions in both continuous and discrete network design problems [55] [56]. Fuzzy Logic Models (FLMs), already discussed in the article, also provide solutions for shortest path optimization [57] [58]. They have shown to outperform statistical and probabilistic models like logistic regression. Deep reinforcement learning has demonstrated promise in optimizing traffic control policies embedded in large-scale ITS systems [59]. Agent-Based Software Engineering (ABSE) creates optimal conditions for dynamic, real-time optimization by incorporating multi-criteria decision-making (MCDM) processes [53]. Agentic AI can automate at a higher, more overarching level, more comprehensively, and in a more integrated manner, and can be applied to traffic engineering to solve bigger and more complicated problems.

The intricacies of vehicle accidents vary spatially and temporally. This presents unique opportunities for AI to grab and use as challenges to address by capturing spatial-temporal patterns from accident databases [53]. Hazardous locations can be pinpointed in this way. The development of a Stack Denoise Autoencoder Simulation model showcased an innovative technique for predicting the risk levels of traffic accidents [60]. Unsupervised machine learning models using techniques like K-means clustering and the Apriori algorithm are capable of discerning usable patterns in historical vehicle crash datasets [61].

Shared mobility services have seen successful implementations in transportation through companies like Uber and Airbnb. AI enhances user experiences by personalizing ride suggestions based on historical data and current data, and implementing route-based pricing models that consider various factors such as time of day and location [62]. AI algorithms help in identifying fraudulent activities in shared mobility services [63].

In the public transport sector, hybrid ACOs have been employed for efficient bus scheduling for promulgating demand-responsive transit (DRT) [64]. Deep learning and KNNs have been applied to predict bus arrival times aimed at reducing passenger waiting periods [65]. Innovations like the iBus and Alpha Bus in China demonstrate the potential of automated buses [66], which mimic human driving behaviors through perception, decision-making, and action-taking processes. Other examples of DRT systems, such as Optibus, showcase AI's capacity to adjust routes based on real-time data [67].

## Discussion, Challenges, and Limitations of AI in Traffic Management

The integration of AI-driven traffic systems presents several challenges and limitations. One of the most critical issues is confirming the privacy and security of data. Traffic systems that depend on AI require extensive data on vehicle movements, traffic patterns, and at times personal information from drivers. Handling this sensitive data responsibly is essential to maintain public trust.[68] [69] The risk of cyber-attacks on AI-grounded transportation management systems is a significant threat. Malicious actors like red hats may be able to manipulate traffic signals, disrupt traffic flow, and endanger public safety using cyber-attacks. Robust cybersecurity measures and secure data handling practices are paramount to mitigate these risks [70] [71] [72].

AI bias, privacy concerns and other ethical considerations are also major concerns in the deployment of AI-based traffic systems [73]. The interpretability of AI models in traffic management is another challenge [74] [75] [76]. AI models like machine learning-based models, especially ones within the realm of deep learning, are increasingly tedious to unravel, making it hard to unravel their workings and explain how they arrived to their outcomes. Explainable AI (XAI) techniques like LIME and SHAP can be beneficial in this, and traffic experts and engineers may collaborate with computational scientists to render a more effective paradigm of explainability and interpretability with regards to AI algorithms in traffic management.

AI models are trained on historical data. Historical data reflects existing biases in traffic enforcement and resource allocation. This usually leads to unfair outcomes, inter alia disproportionate traffic enforcement in certain neighborhoods and biased decision-making in traffic control. For public acceptance and ethical governance, it must be ensured that these systems are designed and tested to be fair and unbiased is vital. AI algorithms can perpetuate existing biases if trained on biased data, leading to discriminatory enforcement of road laws. For example, if an AI-oriented system is trained on data from predominantly low-income neighborhoods, it may disproportionately target drivers in those areas. Automated traffic enforcement systems can collect and store large amounts of personal data, raising privacy concerns. This data could be misused or shared with third parties without proper consent. Over-reliance on AI for enforcement leads to complacency and a decline in human surveillance. This can result in errors and unfair treatment of drivers.

There are ethical dilemmas associated with AI decision-making in traffic scenarios [77]. This includes the use of AI for automated traffic enforcement and prioritizing certain types of vehicles over others. Prioritizing emergency vehicles and public transportation may raise issues related to fairness and equity. Prioritizing emergency vehicles may cause the emergency response drivers to overspeed, causing safety risks, including tailbacks and collision mishaps. Prioritizing vehicles with higher emissions may have negative environmental consequences. These ethical challenges must be carefully considered and addressed through transparent policy frameworks and public dialogue.

Infrastructure and cost challenges are another significant barrier to the adoption of AI in traffic management. Implementing AI technologies requires substantial investments in both hardware and software. It also requires investment in ongoing maintenance and upgrades. Many cities, especially those in developing regions, do not possess the necessary financial resources and technical expertise to support such advanced systems. Existing infrastructure needs modernization to accommodate AI technologies, inter alia the installation of smart sensors, upgraded traffic lights, and augmented communication networks. These high costs and logistical challenges usually hinder the adoption of smart transportation systems. This is particularly true for areas with limited budgets and aging infrastructure.

Scalability and adaptability also create a host of challenges. Different cities have distinct traffic patterns, infrastructure layouts, and transportation needs [78]. This makes it difficult to implement a one-size-fits-all solution in the purview of intelligent transportation management. AI models that work well in one municipal area may not be effective in another due to differences in road design, driver behavior, sociocultural aspects, financial considerations, existing infrastructure, and variability in traffic regulations. Developing AI systems that are adaptable and scalable to various urban settings is essential for their successful deployment. This requires creating flexible models that can be customized to local conditions and can be continuously updated based on real-time data and feedback. Stakeholder collaboration and meaningful discussions with actionable outcomes can resolve most of these challenges.

*Challenges Associated with Integrating Autonomous Vehicles in the Current Paradigm*

The Society of Automotive Engineers (SAE) has developed a set of standards defining the levels of autonomy in vehicles. The SAE has set six levels in this classification [79]. These levels range from no automation (Level 0) to full self-driving capabilities (Level 5). Each level outlines the extent of human involvement in driving and the degree to which a vehicle can independently manage functions such as steering, braking, and accelerating.

At Level 0, the driver remains fully responsible for the car, although features like automatic emergency braking, blind spot monitoring, cruise control (which may be turned off by the driver if needed) and lane departure understanding may be present in the vehicle. Level 1 introduces automated driver assistance with features like adaptive cruise control (automatic accelerating and braking), automated lane-keeping, and autonomous reverse parking systems that provide support yet require the driver's attention. Level 2 integrates more advanced systems, like ADAS. These systems let the car handle more driving tasks, but the vehicle yet requires a certain level of intervention from the driver [80].

Level 3 autonomy, termed conditional driving automation, permits the driver to become more of a passive passenger. All the same, the driver must be ready to regain control when requested by the vehicle itself.

Level 4 autonomy, termed high driving automation, involves the car basically becoming a completely self-driving vehicle that does not need human intervention in areas and conditions where high definition (HD) maps are available. Level 5 autonomy, termed full driving automation, does away with the steering wheel and pedals and drives on its own under all circumstances. Equipped with level 5 autonomy, a vehicle can navigate any milieu under any conditions. Such an automotive capability absolutely obviates the need for a human driver.

Many vehicles that operate nowadays sport Level 2 systems. Some of them are produced by the likes of Tesla, GM, and Mercedes. Although Level 3 is still under regulatory review, some automakers, like Honda, have tested vehicles with level 3 autonomy in limited regions. The automotive industry is gradually moving toward vehicles armed with level 4 autonomy. Ongoing development and testing processes are rapidly making driverless driving technology more accessible each year [81].

Tesla leads this trend of churning out AVs by introducing an "Autopilot" system in some of its vehicles. It is a Level 2 feature, according to the SAE. This system assists a driver but still requires human oversight. Within the overall Autopilot feature, Basic Autopilot is an ADAS feature that Traffic-Aware Cruise Control and Autosteer. Traffic-Aware Cruise Control maintains a set speed and a safe following distance from the vehicle ahead, while Autosteer helps keep the vehicle centered in its lane [82].

Enhanced Autopilot adds more features such as Auto Lane Change, which automatically changes lanes when the turn signal is activated and Autosteer is engaged. "Navigate on Autopilot" guides the vehicle through highway interchanges, including lane changes and exit selection. Autopark helps the vehicle engage in parallel and perpendicular parking. Summon is a functionality which allows the Tesla vehicle to move forward and backward remotely. Smart Summon enables the vehicle to navigate around obstacles like other vehicles and maneuver itself in intricate backdrops to meet the driver or reach a designated location.

Full Self-Driving Capability is the most advanced level of driver assistance offered by Tesla within its Autopilot suite. It includes Traffic Light & Stop Sign Control, which helps the vehicle slow down and stop for traffic signals and stop signs. Autosteer on City Streets, also referred to as Full Self-Driving (Supervised) by Tesla, has the capacity to navigate city streets. It can empower a vehicle to make turns, enter and exit highways correctly, and handle intersections.

Audi has introduced "Traffic Jam Pilot" on its A8 Saloon. This offers level 3 autonomy. It allows hands-off and eyes-off driving in areas afflicted with traffic snarls. Even so, due to legal hurdles, Audi has not activated this system in its vehicles available in the market. Companies like Waymo and Cruise have been making progress with AVs. They have been testing new autonomous systems. Waymo has covered over twenty million miles on public roads with its AVs and Cruise has obtained permission to test cars furnished with level 4 autonomy in the State of California.

Vehicles with the right hardware can be updated through software over time. This integrates features such as navigation and infotainment. Automakers can potentially monetize these technologies through software-as-a-service and deliver updates over-the-air (OTA) [83]. Today's vehicles already offer features like automatic emergency braking and lane-keeping assist. These ADAS will evolve into fully autonomous systems, enabling functions like "automated valet parking" and "smart summoning," where a vehicle can navigate independently to a parking spot and to a pickup point. These technologies pave the way for future robotaxi services.

Some potential functions of AVs have been categorized and highlighted by Stanley and Gyimesi [53] [84]. These comprise self-healing, self-socializing, self-learning, self-driving, self-configuring, and self-integrating capabilities. Self-healing is self-explanatory and makes AVs capable of diagnosing and fixing their internal issues. Self-socializing enables AVs to safely interact with infrastructure, other vehicles, and humans, including the ability to contribute to V2X communications, without posing risks. Self-learning includes the concept of continuous learning as studied in machine learning, and the concept of observational learning as illustrated in the field of robotics. Self-learning ensures AVs continuously improve by adapting to environmental and user inputs. Self-driving is also self-explanatory and focuses on autonomous navigation, which is the chief aim with which AVs have been conceived. Avs are indeed synonymously called self-driving vehicles. Self-configuring is a type of self-learning capability which helps an AV personalize the experience it renders to its user based on the user's preferences. Self-integrating or self-integration allows an AV to connect seamlessly with broader transportation systems, without presenting any risk. Efficient self-integration is one of the primary intended functions that an AV must embody.

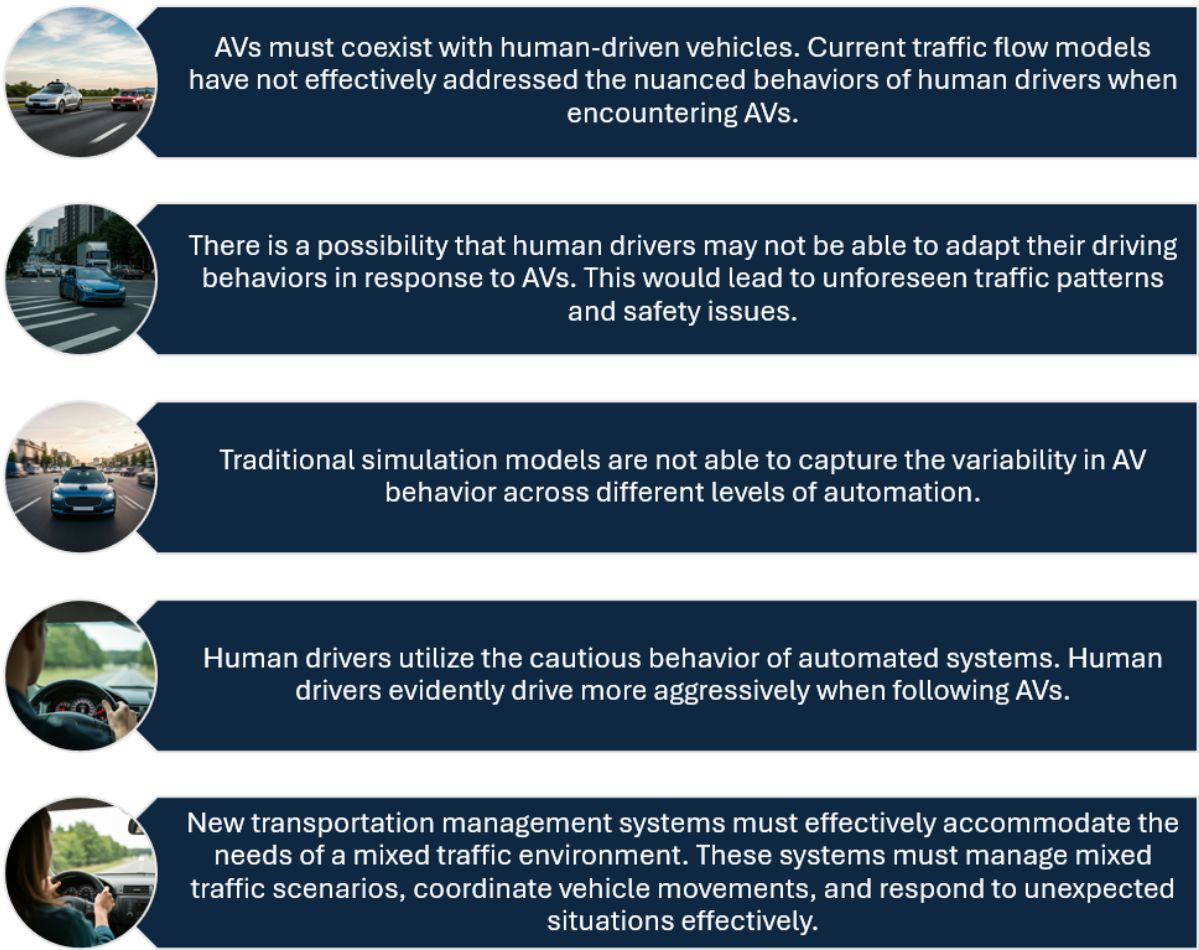

Figure 6: Challenges Associated with AVs and Human Drivers Sharing On-road Space

The integration of AVs into existing traffic systems presents both opportunities and challenges. Self-driving cars have the prospect of significantly magnifying traffic flow and safety. However, they must coexist with human-driven vehicles. This might lead to erratic results. AI plays a critical role in managing these interactions. AI ensures that AVs can navigate well, and human drivers can be safe in the presence of AVs. Managing mixed traffic scenarios, coordinating vehicle movements, and responding to unexpected situations, are essential for the successful merging of AVs into the broader traffic ecosystem [52] [77] [85] [86].

The current traffic flow models which simulate interactions between AVs and human drivers have been inadequate in addressing the nuanced behaviors of human drivers when they encounter AVs on the road [87]. The gap in understanding was largely due to a lack of available data in the field, resulting in this aspect remaining largely unexplored. It has been suggested that human drivers are more likely to adapt their driving behaviors in response to the predictable and cautious nature of AVs. This can perchance lead to unforeseen traffic patterns and safety issues [88]. Drivers' behavior is a major factor in road accidents and a significant contributor to insurance claims. This makes understanding human-AV interactions substantially important as AVs continue to integrate into the current traffic ecosystem [89]. One of the complexities not captured by traditional simulation models is the variability in AV behavior across different levels of automation, from driver assistance to full autonomy. In mixed-control situations, the behavior of AVs as well as that of human drivers can potentially become more unpredictable [90].

Studies suggest that human drivers are learning to adjust their driving strategies in response to AVs. Human drivers utilize the cautious behavior of automated systems [91]. They often mimic the cautious behavior of these automated systems but exhibit more belligerent driving when following them. This shift in behavior can heighten the risk of collisions and negatively impact traffic flow. For instance, data from Waymo shows that human drivers tend to drive more aggressively when trailing a Waymo driverless car [92]. This would dilute the intended benefits of instituting AVs on the roads. Human-driven vehicles and AVs are expected to share the roads for the foreseeable future. In this context, it becomes imperative to develop transportation management systems that accommodate the needs of such a mixed traffic environment [93] [94].

Some studies also reveal that human drivers may interact differently with AVs than with human-driven vehicles. These drivers exploit AVs' risk-averse behavior. One study found that while participants revealed no significant differences in safety margins when turning in front of AVs and human-driven vehicles, they demonstrated greater confidence in their decision-making around AVs. This was indicated by reduced variance in gap size acceptance and distinct brain activation patterns [95].

With regards to interaction with pedestrians, just the same, AVs have showcased better empathy as their design is inherently targeted towards the safety of pedestrians they may encounter on the road. On a test drive, a driverless AV in California stopped for a pedestrian who had started crossing the road at a time and possibly at a spot where they were not supposed to cross (jaywalk). The on-road vehicles had the right of way in that situation, and a human-driven vehicle utilized this clause, almost endangering the pedestrian's life. Yet, the AV stopped for the pedestrian. AVs have also evinced considerable potential in bringing down the frequency of road accidents, including rear-ended, T-bone and frontal collisions.

Remote-driven vehicles are another category of AVs. With drives on long highways over several hours to days being dreaded for their tedium and potential to cause immense exhaustion in drivers of large cargo vehicles like trucks and semi-trucks, remoted-operated vehicle technology is gaining traction. This will allow the driver to sit in front of a laptop or a personal computer screen at home and drive freight over thousands of miles, back and forth, without themselves moving a mile.

Apart from societal acceptance, challenges for AVs also span areas such as data recognition and software integrity. Data recognition is a core aspect of autonomous driving systems that rely on sensor technologies like LiDAR, radar, and cameras to interpret the environment. Currently, these systems struggle in adverse weather conditions like when it snows or rains, poor lighting, and unexpected, unusual and complex urban scenarios and rare driving conditions. AVs misinterpreting road signs, failing to detect pedestrians, and responding inaccurately to sudden changes in traffic can pose grave risks to proximal vehicles, pedestrians, and infrastructure. Deep learning models that drive AV perception require massive datasets for training and validation. The variability of real-world conditions makes perfect recognition an ongoing challenge.

Furthermore, edge cases (unusual driving situations such as encountering an overturned semi which carried petroleum or hazardous or inflammable material) remain a concern, because they are harder to predict and solve algorithmically.

Autonomous systems are highly dependent on complex software stacks, which, for road safety and effective driving, must operate reliably and consistently under all conditions. Software malfunctions due to coding errors, sudden automatic system updates, or owing to cyber-attack vulnerabilities can have disastrous consequences in real-world applications of AVs. We must ensure that AVs meet rigorous safety standards and are resilient to external interferences like hacking and system glitches. This requires conscious endeavors in applying software verification techniques and cybersecurity measures to AV systems. As AVs increasingly become participants in V2X communication infrastructure, maintaining the integrity and security of the data exchanged between AVs and other entities is essential to preventing traffic accidents and system failures. For example, a speeding AV, as a part of V2P communication, must display or transmit warnings to pedestrians to stay safe and as far away as practicable. If it fails to do so, a failure may occur.

An ethical challenge associated with AVs involves decision-making in life-and-death situations. Autonomous vehicles, in certain high-risk scenarios, may have to make split-second decisions about whether to prioritize the safety of their passengers or that of pedestrians, or, for example, two groups of pedestrians, one with a few women and kids and one with a large number of healthy adult males. These decisions bring forth moral and philosophical questions, where an AV must weigh the lives of one group over another. AVs may be intelligent but are not sentient or conscious, and the thinking required for taking such decisions may be beyond their capabilities. The algorithms behind these systems are designed by humans who must program responses to these situations, yet there is no consensus on how to assign moral weight to different outcomes.

Accountability in accidents involving AVs introduces another ethical challenge. When an AV is involved in a crash, determining if the manufacturer, the software developer, or the individual inside the vehicle, if any, is responsible, can be a difficult legal and ethical question. Traditional laws of liability, which rely on human error, do not apply to autonomous systems. With fully autonomous vehicles, the role of the human driver becomes less clear. The decision-making frameworks must be participatory and inclusive and must apportion responsibility fairly.

As with humans learning to drive, the true measure of success for an AV undergoing testing requires testing on public roads with vehicular traffic and pedestrians in the vicinity. This raises the ethical dilemma of using public spaces as testing grounds, where the general population, including pedestrians, cyclists, and other drivers, might be unaware they are sharing the road with vehicles still in development. Also, an AV still being tested might not be fully capable of saving lives on the road that it itself might endanger. This raises ethical questions about real-world AV testing.

*Impact of AI-based Traffic Systems on Environment and Sustainability*

AI-driven traffic management systems have the potential to significantly impact the environment and promote sustainability in urban areas [96] [97] [98]. One of the primary benefits of AI-based transportation management systems is the reduction of traffic logjams and emissions [99] [100]. AI minimizes idle times and causes moderations in stop-and-go driving patterns. Such driving patterns are major contributors to fuel consumption and greenhouse gas emissions. Smart traffic lights, adaptive signal control, and real-time traffic prediction synchronize traffic flow. Such technologies reduce the carbon footprint of urban transportation systems. This leads to cleaner air, helps cities with emission reduction, and mitigates the adverse impacts of climate change [101] [102].

AI-powered traffic systems promote sustainable urban mobility. These systems support alternative modes of transportation such as bike-sharing, EVs, and carpooling systems [103]. AI-enabled platforms provide real-time information on the availability of shared bikes, finetunes routes for EVs considering charging station locations, and match commuters for carpooling opportunities [104]. These technologies make it easier and more convenient for people to choose sustainable transportation options over single-occupancy vehicle use. AI algorithms analyze demand patterns and dynamically allocate resources to areas with high demand for bike-sharing, bicycles and scooters, and carpooling. This eases the accessibility and delivery of these services [105].

Smart transportation systems encourage the adoption of greener mobility options. AI helps reduce the reliance on private vehicles [106]. This supports the development of more sustainable urban transportation ecosystems. Smart traffic systems powered by AI support climate change mitigation strategies [107]. AI nurtures urban resilience. Nevertheless, the successful implementation of these technologies begs careful planning, meticulous investment in infrastructure, and conscientious collaboration among various stakeholders, inter alia governments, technology providers, grassroots organizations and the public.

## Future Directions in AI for Traffic Management Systems

The future of AI in traffic systems is set to be transformative. Its future is primed to be helmed by advancements in sensor fusion, mechatronics, communication technologies, multimodal transportation integration, and ethical AI development. A key area of growth is the integration of AI with 4G and 5G networks and the IoT [108] [109]. The more developed communication capabilities will be, the lower the latency and the higher the data transfer rates will be. Such potential is exhibited by the proliferation of 5G. It is poised to enable real-time data exchange between vehicles, traffic signals, and other infrastructure components at unprecedented speeds and data transfer rates in V2X communications [110] [111]. This, along with advancement in processing, such as parallel processing and distributed computing, will allow AI systems to quickly transmit vast amounts of data from IoT devices, such as smart sensors and connected vehicles, and analyze it [112]. The probability of vehicular queuing will be reduced, emergency services will be more responsive, and incident response will be rapider. As cities adopt more IoT-enabled devices, the synergy between AI, 5G, and IoT are primed to become pivotal in creating smarter, more adaptive, more accountable, more citizen-friendly, and more sustainable urban environments [113].

Autonomous traffic control centers represent another promising direction in the evolution of AI for traffic systems. Such systems will anticipate vehicular tailbacks, dynamically adjust traffic signals, and reroute vehicles in real-time based on current conditions by using predictive analysis [114]. Autonomous traffic management will also handle emergencies and streamline the movement of emergency vehicles [115]. The trend is moving towards systems that can operate independently. This will reduce the need for manual monitoring and intervention. However, human surveillance over such intelligent systems must continue to be essential.

AI's potential to propagate advanced multimodal transport systems is gaining footing. Coordinating diverse transportation modes such as road, rail, and air travel involves complex challenges. AI is uniquely positioned to address these challenges [116] [117]. In future, AI will potentially help adjust schedules, manage interchanges between different modes of transport, and predict demand for various services. This will ensure smoother and more flourishing integration of transportation networks. This intelligent capability is especially important in large urban and metropolitan centers. AI can help cities offer more cohesive transit and user-centered experiences [118].

The ethical considerations surrounding the use of AI will also become important. There is a growing need to develop more transparent and accountable AI systems that can be trusted by the public [119]. Ethical AI in traffic management must ensure that AI algorithms are free from biases, that they make decisions in a fair and equitable manner, and that there are mechanisms in place to explain and audit these decisions. This is where human oversight becomes vital [3]. Addressing these ethical challenges will be essential for gaining public acceptance and trust in AI-driven traffic solutions. This is particularly so in a likely scenario where Ai-based systems become more autonomous and influential in daily urban life.

Research and innovation in AI traffic systems are advancing rapidly. Numerous ongoing initiatives explore new technologies and methodologies. Areas such as deep learning, reinforcement learning, Bayesian networks, logistic regression, semi-supervised and unsupervised learning, evolutionary computing, and quantum computing are being researched for their potential to modernize traffic predictions, to restructure routing, and to augment the success of traffic management systems [120]. Collaborative research efforts between academic institutions, government agencies, non-governmental entities, public, and private companies are driving the development of next-generation AI traffic solutions forward. As more real-world data becomes available and computational capabilities continue to grow, the possibilities for innovation in this field are expanding. This will soon pave the way for smarter, more adaptive, and more efficient traffic systems.

AI and its various methodologies including machine learning and NLP have evoked optimistic prospects in multiple fields [121] [122] [123] [124]. There is no reason why the area of traffic management will remain aloof from its positive influence. In the near future, it is hoped that automation will pervade traffic management to a massive extent. This will simplify and reinvigorate the on-road experience of vehicle-users as well as pedestrians. The penetration of AI will make the roads a panacea for travelers, commuters and tourists alike. Under the aegis of responsible, ethical, and controlled application of AI and its subfields, it is hoped and forecasted that they will encounter safer, more sustainable, and more enjoyable transportation experiences.

## Conclusion and Future Work

The integration of AI in traffic management shows credible promise to revolutionize transportation management. With the incorporation of AI, urban transportation is becoming more responsive to the needs of their citizens. AI technologies are briskly advancing and converging with emerging trends, inter alia, self-driving vehicles, V2X communication protocols, sensor fusion, and autonomous systems. This makes the possibility for innovation in traffic management vast and optimistic. The capability of AI-powered systems to analyze vast amounts of real-time data to make valid predictions about traffic patterns, which improves traffic movement and enables traffic administration to augment its measures to reduce on-road overcrowding, to improve air quality, and to enhance the experience for all the stakeholders. The current drive is towards seamless integration of AI with existing transport infrastructure. When coupled with human supervision with conscientious respect for ethical considerations, such steps will ensure that AI-fueled transportation systems are not only efficient but are also transparent, accountable, and trustworthy. This is going to spread a red carpet for a futurity where transportation is safer, more enjoyable, and more sustainable for every user. In this way, smart traffic planning will stimulate a future where cities are smarter and more sustainable, where road travel is smooth and uncongested, and where drivers, passengers, pedestrians and residents are safe and are able to enjoy a sustainable and healthy environment.

*Declaration of Funding*
The author(s) did not receive any funding for this article.